\documentclass[sigconf]{acmart}

\usepackage{booktabs}
\usepackage{xcolor}
\usepackage{amsmath}
\usepackage{pgfplots}
\usepackage{etoolbox}
\usepackage{multirow}
\usepackage{float}
\pgfplotsset{compat=1.17}

\renewcommand\footnotetextcopyrightpermission[1]{}
\renewcommand\acmConference[3]{}
\renewcommand\acmYear[1]{}
\acmYear{}
\acmDOI{}
\acmISBN{}
\begin{document}

\title{From Gradient-Boosted Trees to Deep Recommenders: Practical Lessons from
  Migrating a Production Customer Support Recommender}

\author{Sonia Sharma}
\affiliation{\institution{Intuit}\country{USA}}

\author{Jeyendran Balakrishnan}
\affiliation{\institution{Intuit}\country{USA}}

\author{Shreya Rajpal}
\affiliation{\institution{Intuit}\country{USA}}

\author{Swapnil Parekh}
\affiliation{\institution{Intuit}\country{USA}}

\author{Nagaraj Janardhana}
\affiliation{\institution{Intuit}\country{USA}}

\author{Andrew Mattarella-Micke}
\affiliation{\institution{Intuit}\country{USA}}

\renewcommand{\shortauthors}{Sharma et al.}

\begin{abstract}
Product catalogs in fast-moving service businesses are shifting from static, independently priced SKUs toward dynamically bundled, discount-coupled offerings—a shift that strains the tree-based classifiers traditionally preferred for sparse and highly imbalanced data. These classifiers assume a fixed, slowly changing label space and struggle to incorporate multimodal signals such as tabular data and transcripts.

We present the migration of a live, production conversational recommendation system from a gradient-boosted multiclass model to a pairwise-binary deep recommender. Because this system is critical to ecosystem growth initiatives and downstream features like dynamic pitching --- surfacing the most relevant pitch text to a support agent in real time during a live customer conversation --- maintaining live recommendation quality was a non-negotiable constraint.

We detail the techniques that made this migration successful: reformulating recommendation as pairwise binary prediction to learn jointly from user and item features, and enhancing learned representations via negative sampling and noise injection. To efficiently incorporate long, live conversation context, we apply attention pooling over transcript chunks and benchmark it against TF-IDF and sentence-embedding baselines. Finally, we explore multiple architectures (including two-tower models, DeepFM, and their variants) and loss functions such as contrastive loss. Evaluating against a CatBoost baseline across all conversational stages, we demonstrate that our approach achieves parity at conversation beginning and outperforms at later conversational stages.
\end{abstract}


\maketitle
\thispagestyle{empty}
\pagestyle{empty}
\section{Introduction}
\label{sec:intro}

\subsection{The catalog is changing faster than the model can adapt}
\label{sec:why-catalog}

Service businesses are moving from independently-priced, SKU-level offerings
toward dynamically bundled products, often coupled with discounts assembled
at serve-time: a shift in the \emph{shape} of the recommendation problem,
not just its scale, since a bundle's identity is compositional (which SKUs,
which discount) rather than atomic.

The system we are replacing --- a per-product gradient-boosted multiclass
classifier --- was built for the slower-moving catalog. Three structural
limitations motivate the migration:

\begin{enumerate}
  \item \textbf{Problem formulation.} Multiclass-over-products assumes a
    fixed label set; adding a bundle means adding a class and retraining
    the label space itself. A \emph{pairwise-binary formulation}
    $(\mathrm{contact}, \mathrm{product}) \rightarrow
    \mathrm{match}/\mathrm{no\text{-}match}$ sidesteps this: a bundle or
    new SKU is just a new item-side feature vector, not a new output unit.
  \item \textbf{Multimodal signal.} The legacy model scores off tabular,
    aggregated features and cannot natively consume the live conversation
    transcript. Deep recommenders can condition directly on transcript
    representations via attention pooling, which learns a per-chunk
    relevance weight instead of pooling the conversation uniformly
    (Section~\ref{sec:text}).
  \item \textbf{Architectural and objective flexibility.} Once the problem
    is pairwise, contrastive objectives over a shared user/item embedding
    space become available (Section~\ref{sec:contrastive-loss}), as do
    architecture families (two-tower, DeepFM; Section~\ref{sec:arch}) that
    trade off differently on latency and expressivity.
\end{enumerate}

\subsection{Costs of the migration}

Moving to a deep recommender introduces three concrete costs: a larger
tuning surface (an order of magnitude more hyperparameters than a
boosted-tree model, addressed in Section~\ref{sec:negsampling} and the
double-descent scan in Section~\ref{sec:dd}); a sparser, more imbalanced
supervision signal (positive rate $\approx 4.5\%$, 154 categorical fields,
$\sim$2k dense features, addressed by the same two sections); and a higher
per-request inference cost, since every contact-product pair now needs a
forward pass through embedding lookups and a DNN rather than one
boosted-tree traversal (Section~\ref{sec:inference}).

\subsection{Contribution}

This paper's contribution is the set of concrete techniques required to
carry out this migration without regressing recommendation quality,
evaluated end-to-end on live customer-support recommendation data rather
than benchmark data:

\begin{itemize}
  \item A pairwise-binary problem reformulation (Section~\ref{sec:setup})
    that admits arbitrary item/context features without requiring a fixed
    label space.
  \item A negative-sampling study (Section~\ref{sec:negsampling}) and why we
    settled on the mixture we did.
  \item A contrastive objective (Section~\ref{sec:contrastive-loss}) over the
    shared user/item embedding space.
  \item Attention pooling over conversation chunks
    (Section~\ref{sec:text}) to identify which part of a long call the
    model should condition on, compared against TF-IDF and dense
    sentence-embedding baselines.
  \item A double-descent characterization (Section~\ref{sec:dd}) of
    DeepFM's capacity/regularization regime --- explaining
    \emph{why} its overfitting training curve is not the
    dead end it looks like, and what capacity range actually helps.
  \item An architecture comparison (Section~\ref{sec:arch}) between
    two-tower and DeepFM formulations.
  \item Inference-time optimizations (Section~\ref{sec:inference}) required
    to serve the resulting model within the customer-support system's
    real-time latency budget.
\end{itemize}

Throughout, we evaluate at three deployment-relevant points in a live
conversation --- \textbf{Case-Load (CL)}, \textbf{Early-Recommendation
(ER)}, and \textbf{Ground-Zero (GZ)} --- defined precisely in
Section~\ref{sec:metrics}, because the production constraint is not ``good
average ranking quality'' but ``good enough recommendation quality at the
specific moment the downstream system consumes it''
(Section~\ref{sec:constraint}).

\subsection{The constraint: recommendation quality cannot regress}
\label{sec:constraint}

This migration was undertaken under a hard non-regression constraint, not a
green-field research question. Our customer support recommender feeds a
downstream pitch/talk-track augmentation system, DynaPitch~\cite{dynapitch},
whose own quality depends directly on the quality of the recommendations it
augments. A regression here is not an isolated metric drop; it propagates
into that system's output.

More broadly, this recommender is the backbone several ecosystem growth
and retention initiatives are built on, feeding upsell recommendations
that support a revenue portfolio on the order of \$360 million, which is
why quality preservation, not just improvement, is the constraint this
paper is written against. Every experiment is therefore reported against
the deployed CatBoost baseline on the \emph{same} fixed positive-contact
population, using the \emph{same} metric definitions, so ``did we
regress'' has an unambiguous answer at every step
(Section~\ref{sec:metrics}).

\section{Problem Setup and Evaluation Methodology}
\label{sec:setup}

\subsection{Real-time recommendation setup}

A customer-support conversation is scored incrementally as it progresses; a
recommendation is served at some point in the call. The production system
chunks the transcript into \texttt{chunk\_num}-indexed segments of exactly
20 utterances, with one scoring pass per chunk --- Case-Load and Early-Reco
are chunks 0 and 1, the first and second 20-utterance blocks (``at call
start'' is intuition for chunk 0, not a wall-clock claim). 20 utterances
balances having enough context to score against low per-chunk invocation
cost at serving time.

\subsection{Data and train/valid/test split}

Throughout this paper, a customer-support conversation is referred to as a
\emph{contact}. The train/validation pool spans contacts from 2025-01
through 2026-03; within that pool, contacts are randomly split 75/25 into
train and validation at the \texttt{contactid} level (seed 42), so every
chunk row of a given contact stays in the same partition. Test is a
temporal, out-of-time holdout: contacts from 2026-04 through 2026-06,
unseen at any point during training or validation. This yields
903{,}829 / 302{,}122 train/valid rows and 360{,}022 test rows after
chunking and negative sampling.

\subsection{Chunk-level and contact-level metrics}

Two levels of evaluation, matching a shared evaluator so every architecture
in this paper is scored identically:

\begin{itemize}
  \item \textbf{Chunk/pair level} --- flat (contact-chunk, product) rows:
    log-loss, ROC-AUC, PR-AUC (average precision), all fit on validation.
  \item \textbf{Contact level} --- one decision per contact at a defined
    point in the call.
\end{itemize}

\subsection{The three business metrics: CL / ER / GZ}
\label{sec:metrics}

\begin{table}[h]
\centering
\caption{Contact-level business metrics.}
\label{tab:metrics-defs}
\small
\setlength{\tabcolsep}{3pt}
\begin{tabular}{@{}p{0.13\linewidth}p{0.26\linewidth}p{0.46\linewidth}@{}}
\toprule
Metric & Selection & What it answers \\
\midrule
CL (Case-Load) & first chunk (\texttt{chunk\_num}$=0$) & Quality of the earliest possible recommendation \\
ER (Early-Reco) & second chunk (\texttt{chunk\_num}$=1$) & Quality shortly after, with a little more signal \\
GZ (Ground-Zero) & per-contact max-weight chunk & Quality at the point a recommendation is most likely to fire \\
\bottomrule
\end{tabular}
\end{table}

All three are micro-F1 computed over the \emph{same fixed positive-contact
population} used to evaluate the legacy CatBoost model (16{,}709 contacts
in the current data snapshot), so every number in this paper is directly
comparable to the deployed baseline with no population mismatch.

\textbf{Baseline reference values used throughout this paper} (the legacy
model (CatBoost) --- see Section~\ref{sec:formulation} for the exact
CatBoost configuration and negative-sampling scheme this is): CL~0.4985,
ER~0.5310, GZ~0.5676.

\section{Baseline Setup: Negative Sampling and the CatBoost/DeepFM Baseline}
\label{sec:baseline}

This section establishes the vocabulary and negative-sampling setup shared
by every model in this paper --- CatBoost and DeepFM alike --- and reports
both models' baseline numbers under it. The deeper ablation study of
negative-sampling variants is deferred to Section~\ref{sec:negsampling}.

\subsection{A taxonomy of negatives}
\label{sec:neg-taxonomy}

The \emph{choice of negatives} is the dominant modeling lever on this
problem, consistent with the broader implicit-feedback recommendation
literature~\cite{rendle2009bpr}: severe pair-level imbalance and
invisible non-converted candidates can otherwise distort the model's
per-contact top pick toward over-represented products. We use one
vocabulary for negative types throughout this paper, defined at two
levels:

\begin{itemize}
  \item \textbf{Explicit vs.\ implicit} (top-level split, by \emph{how} we
    know a candidate is negative). \textbf{Explicit} --- the customer gave an
    explicit answer: they said no to this specific product; this is the
    \texttt{product\_\allowbreak explicit\_\allowbreak negative} pipeline
    mechanism. \textbf{Implicit} --- no explicit rejection was ever given;
    all we know is the contact did not convert on this candidate.
  \item \textbf{Hard vs.\ soft} (within each of explicit and implicit, by
    \emph{how strong} the negative signal is). On the explicit side, hard =
    an unambiguous, immediate no; soft = a weaker or more ambiguous
    non-acceptance. On the implicit side, hard = the candidate came from a
    chunk where the contact converted on \emph{nothing at all} --- the
    strongest available "no signal" even without an explicit rejection,
    implemented as the \texttt{overall\_negative} and
    \texttt{product\_\allowbreak random\_\allowbreak negative} mechanisms;
    soft = the candidate co-occurred in the same chunk as a \emph{different}
    product the contact did convert on, so its non-conversion is weaker
    evidence (it may simply not have been the one chosen, not actively
    rejected), implemented as the \texttt{implicit\_negative} mechanism.
\end{itemize}

A third, separate concept is the \textbf{pure-negative contact}: a contact
where nothing converted at all, across the whole contact. This is a
population-level property of a contact, distinct from the per-row
explicit/implicit/hard/soft taxonomy above, which describes individual
negative \emph{rows}.

At the contact level, we start from an approximately 1/3--2/3
positive/negative sample (every positive contact kept, negative contacts
sampled to roughly match, explicit-hard first then explicit-soft) --- a
deliberately balanced, not population-representative, sample. Training
rows, however, are contact-\emph{chunk} samples, not contacts: negative
contacts run longer than positive ones (they keep generating chunks
instead of ending on conversion), and we additionally construct implicit
negatives for every positive contact-chunk, so a contact-level sample
built to be roughly 1/3 positive becomes a chunk-level training
population that is only $\approx$4.1\% positive.

\subsection{Chunk-level vs.\ contact-level sampling granularity}
\label{sec:chunk-vs-contact}

Implicit-hard negatives (\texttt{overall\_negative} and
\texttt{product\_\allowbreak random\_\allowbreak negative}) can be sampled at two granularities:
once per \emph{contact} (the same negative set stamped on every chunk) or
independently per \emph{chunk}. This finding applies identically to
CatBoost and DeepFM, since it concerns how training rows are constructed,
not which model consumes them. At matched $K{=}2$, sampling independently
per chunk beats sampling once per contact on both reported cuts (CL 0.437
vs.\ 0.407, ER 0.530 vs.\ 0.525; GZ was not evaluated for the
contact-level run) --- the convention we adopt for the rest of this
paper.

\subsection{Establishing the baselines}
\label{sec:formulation}

\textbf{Scope note --- pairwise-binary formulation throughout.} This
paper does not compare against an earlier multiclass formulation. Both
models under study here --- the legacy model (CatBoost) and the DeepFM
candidate --- already run the same pairwise-binary formulation on the same
data: each row is a (contact-chunk, candidate product) pair with a binary
match/no-match label, and negative rows constructed per the taxonomy in
Section~\ref{sec:neg-taxonomy}. Section~\ref{sec:why-catalog}'s multiclass
discussion is background motivation only, not a comparison this paper
reports results against.

\textbf{Text features.} CatBoost represents transcript text with a single
TF-IDF vectorizer over chunk text (max 3500 features, 1--3-grams,
stopwords removed, terms in 10--75\% of documents). DeepFM's with-text
configuration TF-IDF-vectorizes the same text, then compresses it via
truncated SVD to $\sim$40 components concatenated into the tabular dense
vector (1980$\to$2020 features) rather than a separate learned text
tower; richer text representations are explored later
(Section~\ref{sec:text}).

\begin{table}[h]
\centering
\caption{Baseline results (pairwise-binary F1).}
\label{tab:formulation}
\small
\setlength{\tabcolsep}{4pt}
\begin{tabular}{@{}lrrr@{}}
\toprule
Configuration & CL & ER & GZ \\
\midrule
Legacy (CatBoost), with text & \textbf{0.4985} & \textbf{0.5310} & \textbf{0.5676} \\
DeepFM, no text & 0.3426 & 0.4362 & 0.4293 \\
DeepFM, with text & 0.4678 & 0.5283 & 0.5158 \\
\bottomrule
\end{tabular}
\end{table}
 
\textbf{DeepFM architecture.} The DeepFM~\cite{guo2017deepfm} backbone
used throughout is a shared DNN tower $(2048,1024,512,256)$, learning
rate $3\times10^{-4}$, weight decay $10^{-4}$, 10 warm-up epochs. Tabular
features split across two paths feeding the factorization-machine (FM)
layer~\cite{rendle2010fm}: a dense projection over $\sim$1980 continuous
features, and a bin-and-embed sparse path (top-100 categorical features
by mutual information~\cite{battiti1994mi}, 50 quantile bins, embedding
dim 32).
The legacy model (CatBoost) baseline uses depth 10, 1000 iterations,
learning rate 0.0406, $\ell_2$ leaf regularization 8.51, and
\texttt{scale\_\allowbreak pos\_\allowbreak weight} 1.003, selected via 25-evaluation Bayesian search.
Both baselines in Table~\ref{tab:formulation} share the same
negative-sampling configuration --- \texttt{overall\_\allowbreak negative\_n=1}
and \texttt{product\_\allowbreak random\_\allowbreak negative\_k=2}, both at contact granularity,
combined with a two-dimensional row-weighting scheme: each negative row's
final weight is the product of a weight keyed by \emph{negative type}
(\texttt{overall\_\allowbreak negative}, \texttt{product\_\allowbreak
explicit\_\allowbreak negative},
\texttt{product\_\allowbreak random\_\allowbreak negative} each get their own weight) and a
separate weight keyed by \emph{product} (each of the nine products has its
own weight, since some are rarer than others and need more emphasis) ---
two independent lookup tables multiplied together, not one flat weight per
row: the shared substrate every model in this paper is trained and
evaluated on.

\section{Architecture Comparison: DeepFM vs.\ Two-Tower}
\label{sec:arch-early}

Before any additional feature engineering, attention pooling, or contrastive
objective is layered on, we ask a narrower question: holding text handling
to each architecture's own standard mechanism, which base architecture is
stronger? Two-tower architectures~\cite{huang2013dssm,covington2016youtube}
use separate
user/context and item encoders with a dot-product or learned similarity at
the top --- well-suited to retrieval-style serving, weaker at modeling
explicit feature interactions. DeepFM \cite{guo2017deepfm} uses a joint
factorization-machine, DNN, and wide component over a shared feature space
--- richer interaction modeling, less naturally suited to
nearest-neighbor-style serving at inference time.

Two-tower's retrieval-style serving advantage is designed for item catalogs
too large to score exhaustively --- not a constraint we currently face
(nine products today, likely low hundreds with planned expansion, well
within DeepFM's exhaustive-scoring range), so that advantage is latent,
not yet load-bearing; the comparison below measures which architecture
wins \emph{on our data and at our current scale}, not which is better in
general.

\begin{table}[h]
\centering
\caption{DeepFM vs.\ two-tower, both with text.}
\label{tab:arch-early}
\small
\setlength{\tabcolsep}{4pt}
\begin{tabular}{@{}lrrrr@{}}
\toprule
Architecture & CL & ER & GZ & Top-1 \\
\midrule
DeepFM (with text) & \textbf{0.4678} & 0.5283 & 0.5158 & --- \\
TT-base: residual (with text) & 0.4575 & \textbf{0.5537} & \textbf{0.6152} & 0.5822 \\
\bottomrule
\end{tabular}
\end{table}

DeepFM wins at and near call start (CL), while the two-tower model is
stronger once more of the conversation has been observed (ER, GZ). We take
this as evidence that DeepFM is the better base architecture in the regime
that matters most under this paper's non-regression constraint --- the
earliest possible recommendation, before there is any transcript signal to lean into ---
and adopt it as the primary architecture for the rest of this paper, while
returning to the two-tower line of work on its own terms in
Section~\ref{sec:twotower}.

\section{Negative Sampling: A Thorough Investigation}
\label{sec:negsampling}

This section reports DeepFM's own negative-sampling experiments, building
on the shared substrate established in Section~\ref{sec:baseline}: a
structured, rule-based alternative to random sampling, and a sweep over
how many random implicit negatives to sample per positive contact-chunk.

\subsection{Structured negative sampling}
\label{sec:structured-negative-sampling}

Beyond varying how many random implicit negatives are sampled, we also
tested a structured, rule-based sampler that selects implicit negatives
through interpretable patterns in the observed customer and product
data, rather than drawing them uniformly at random --- so the sampling
decision can be inspected independently of the neural model. We compare
it against random sampling at $K{=}2$ and $K{=}4$.

\begin{table}[t]
\centering
\caption{Random vs.\ structured negative sampling, two $K$ values.}
\label{tab:structured-negative-sampling}
\small
\setlength{\tabcolsep}{3.5pt}
\begin{tabular}{@{}llrrrrr@{}}
\toprule
$K$ & Sampling & CL & ER & GZ & Val PR-AUC & Test PR-AUC \\
\midrule
2 & Random     & 0.4649 & 0.4924 & 0.4797 & 0.6845 & 0.5669 \\
2 & Structured & 0.4572 & 0.4987 & 0.4836 & \textbf{0.7137} & 0.5527 \\
4 & Random     & 0.4857 & 0.5084 & 0.4982 & 0.5904 & 0.5244 \\
4 & Structured & 0.4495 & 0.5069 & 0.4900 & 0.6722 & \textbf{0.6293} \\
\bottomrule
\end{tabular}
\end{table}

Structured sampling improves pairwise ranking behavior in several
settings, but these gains do not translate uniformly to the
contact-level metrics: at $K{=}4$, test PR-AUC rises substantially
(0.5244$\to$0.6293) while CL actually drops (0.4857$\to$0.4495); at
$K{=}2$, it improves validation PR-AUC with only modest ER/GZ movement
and no CL or test-PR-AUC gain. PR-AUC measures pairwise discrimination
between positive and negative candidates, while CL/ER/GZ measure whether
the correct product actually rises to the top for a positive contact ---
more targeted negatives can sharpen pairwise separation without
improving the final recommendation decision.

Table~\ref{tab:implicit_random_negatives} isolates the number of implicit
negatives sampled per positive contact-chunk, holding architecture and
training objective fixed, to find the $K$ that best trades off ranking
quality against training-set size. To keep this sweep fast, each
configuration was trained on a 50\% subsample of the training data rather
than the full set.

\begin{table}[t]
\centering
\caption{Implicit random negatives, $K{=}0,\ldots,7$, 50\% training
  subsample. DNN $(2048,1024,512,256)$, dropout $0.50$/$0.60$, LR
  $3\times10^{-4}$, weight decay $10^{-4}$, 10 warm-up epochs, batch 512.}
\label{tab:implicit_random_negatives}
\small
\setlength{\tabcolsep}{4pt}
\begin{tabular}{@{}lrrrrr@{}}
\toprule
$K$ & CL & ER & GZ & Val PR-AUC & Test PR-AUC \\
\midrule
0 & 0.4715 & 0.5072 & 0.4901 & 0.7727 & 0.5832 \\
1 & 0.4706 & 0.5119 & 0.4932 & 0.7250 & 0.5810 \\
2 & 0.4649 & 0.4924 & 0.4797 & 0.6845 & 0.5669 \\

3 & 0.4680 & 0.5021 & 0.4906 & 0.6548 & 0.5614 \\
4 & 0.4857 & 0.5084 & 0.4982 & 0.5904 & 0.5244 \\
5 & 0.4405 & 0.5100 & 0.4994 & 0.6274 & 0.5832 \\
6 & 0.4677 & 0.5132 & 0.4978 & 0.5834 & 0.5883 \\
7 & 0.4629 & 0.5007 & 0.4897 & 0.5762 & 0.5805 \\
\bottomrule
\end{tabular}
\end{table}

One takeaway from this sweep: the gap between validation and test PR-AUC
shrinks monotonically as $K$ increases (from 0.19 at $K{=}0$ to near zero
by $K{=}6$--$7$), but this does not translate into a reliable improvement
on the metrics we actually care about --- CL, ER, and GZ move
non-monotonically with $K$ and do not track the shrinking val-test gap.
Given this, $K$ in the range 2--4 is a reasonable choice to work with
going forward, and is what we use throughout the rest of this paper.

\section{Contrastive and Binary Cross Entropy Loss Training}
\label{sec:contrastive-loss}

Our data contains relatively few positive conversion signals, so we want the model to learn them explicitly and contrast them against meaningful negatives while still accounting for the much larger number of negative examples overall. We therefore use BCE and contrastive loss jointly: BCE captures the overall conversion distribution while contrastive learning sharpens the model's ability to distinguish positive signals from competing negatives.

For each contact-chunk $u$ containing a true positive product $p$, we construct a contrastive group

\[
G_u = \{p,n_1,\ldots,n_K\},
\]

where $u$ denotes the contact-chunk, $p$ denotes a product associated with a true positive conversion, $n_j$ denotes the $j$-th implicit negative product, and $K$ denotes the number of implicit negatives sampled from valid non-positive products. In our best-performing configuration, we use $K=2$ and sample the implicit negatives randomly.

All positive and negative samples in the constructed minibatch contribute to the BCE objective. The BCE loss is defined as

\[
\mathcal{L}_{\mathrm{BCE}}
=
-\frac{1}{N}
\sum_{i=1}^{N}
\left[
y_i \log \hat{y}_i
+
(1-y_i)\log(1-\hat{y}_i)
\right],
\]

where $N$ is the total number of samples in the minibatch, $y_i \in \{0,1\}$ denotes the ground-truth conversion label for sample $i$, and $\hat{y}_i$ denotes the corresponding predicted probability. Positive products have $y_i=1$, whereas implicit negatives and samples from pure-negative contacts have $y_i=0$.

In addition to BCE, we compute a user--item contrastive objective for groups containing at least one true positive product. Let $\mathbf{z}_u$ denote the contact-chunk representation produced by the shared DNN after masking product-specific input features. This masking prevents the contact representation from directly observing which candidate product is currently being scored. Let $\mathbf{z}_p$ denote the representation of product $p$, obtained from its product-text embedding and projected into the same latent space as $\mathbf{z}_u$.

The compatibility between a contact-chunk and a candidate product is
measured using cosine similarity, $s(u,p) = \cos(\mathbf{z}_u,\mathbf{z}_p)$.

For a positive product $p$ and its $K$ implicit negatives $\{n_1,\ldots,n_K\}$, the user--item contrastive loss is an InfoNCE-style~\cite{oord2018infonce,chen2020simple} objective, defined as

\[
\mathcal{L}_{\mathrm{CL}}
=
-\log
\frac{
\exp\left(s(u,p)/\tau\right)
}{
\exp\left(s(u,p)/\tau\right)
+
\sum_{j=1}^{K}
\exp\left(s(u,n_j)/\tau\right)
},
\]

where $s(u,n_j)$ denotes the cosine similarity between contact-chunk $u$ and implicit negative product $n_j$, and $\tau$ denotes the contrastive temperature controlling the sharpness of the relative similarity distribution. In our experiments, the contrastive temperature is fixed to $\tau=1.0$.

Pure-negative contacts do not contain a positive anchor and therefore do not contribute to $\mathcal{L}_{\mathrm{CL}}$. They nevertheless remain part of training through the BCE objective. The contrastive-loss weight is fixed to $1.0$, giving the combined training objective

\[
\mathcal{L}
=
\mathcal{L}_{\mathrm{BCE}}
+
\mathcal{L}_{\mathrm{CL}}.
\]

\subsection{Batch Composition}

Our training data naturally contains approximately one-third positive-containing examples and two-thirds negative examples. Under standard BCE training with random batching, minibatches approximately reflect this underlying $1/3$--$2/3$ distribution in expectation. In the mixed BCE--contrastive setting, we additionally control how positive contrastive groups and pure-negative units are composed within a minibatch.

Let $r$ denote the target fraction of positive contrastive groups used
during batch construction, so a sampling unit is a positive contrastive
group with probability $r$ and a pure-negative unit with probability
$1-r$. We consider $r=\frac{1}{3}$, approximately the natural
positive-group to pure-negative composition of the data, and
$r=\frac{1}{2}$ (50:50), which increases the representation of
positive-containing groups during optimization.

This ratio refers to the composition of \emph{sampling units}, not to the
number of positive/negative BCE labels: when $K=2$, a positive group
contributes one positive and two negative samples, all three to BCE and
the group jointly to the contrastive objective; a pure-negative unit
contributes only to BCE.

To isolate how much of the eventual gain comes from batch composition
alone, independent of adding a contrastive term at all, we compare
$r=\frac{1}{3}$ (random batching) against $r=\frac{1}{2}$ (50:50 group
batching) at $K=3$, training under BCE only in both cases:

\begin{table*}[t]
\centering
\caption{\textbf{Left}: group-balanced batching at $K{=}3$, BCE only, no contrastive term.
  \textbf{Right}: contrastive loss alone vs.\ combined BCE-batching + contrastive loss, at two implicit-negative counts.}
\begin{minipage}{0.47\linewidth}
\centering
\small
\setlength{\tabcolsep}{3pt}
\begin{tabular}{@{}lrrr@{}}
\toprule
Config. & CL & ER & GZ \\
\midrule
Random ($r{=}1/3$) & 0.4680 & 0.5021 & 0.4906 \\
50:50 ($r{=}1/2$) & \textbf{0.4781} & \textbf{0.5441} & \textbf{0.5286} \\
\bottomrule
\end{tabular}
\label{tab:k3-batching}
\end{minipage}%
\hfill
\begin{minipage}{0.51\linewidth}
\centering
\scriptsize
\setlength{\tabcolsep}{2.5pt}
\begin{tabular}{@{}crrrr@{}}
\toprule
$K$ & Setting & CL & ER & GZ \\
\midrule
\multirow{2}{*}{2} & CL only & 0.4636 & 0.4606 & 0.4635 \\
 & BCE+CL & 0.4963 & 0.5433 & 0.5279 \\
\multirow{2}{*}{4} & CL only & 0.4661 & 0.5455 & 0.5048 \\
 & BCE+CL & \textbf{0.4981} & \textbf{0.5510} & \textbf{0.5354} \\
\bottomrule
\end{tabular}
\label{tab:implicit-negative-results}
\end{minipage}
\end{table*}

Controlling minibatch composition alone --- no contrastive term, no
architecture change --- already improves all metrics (ER
0.5021$\to$0.5441, GZ 0.4906$\to$0.5286): an independent, additive lever,
not something the contrastive objective's gains can be fully attributed
to. It is not a substitute either: Table~\ref{tab:implicit-negative-results}
compares contrastive learning alone against the combined BCE-batching and
contrastive-learning objective, at two implicit-negative counts. Adding
the contrastive term on top of 50:50 batching at $K{=}4$ reaches CL
0.4981, ER 0.5510, GZ 0.5354 --- better on every contact-level metric than
batching alone. The strongest contact-level behavior comes from combining
informative negatives, batch composition, and the contrastive objective
together, not any one alone.

\section{Double Descent: Characterizing the Capacity/Regularization Regime}
\label{sec:dd}

\subsection{Motivation}

The DeepFM baseline text model (Section~\ref{sec:text}, 147.8M params)
shows textbook interpolation-onset overfitting: monotone training loss,
validation PR-AUC peaking early (epoch 5, 0.752) then decaying --- the
symptom that leads practitioners to distrust deep models on sparse
tabular data and fall back to trees. Rather than accept this as a
ceiling, we ask whether it is the left half of a double-descent risk
curve~\cite{belkin2019,nakkiran2021} whose right half outperforms both
the DeepFM baseline and the legacy model (CatBoost) it is meant to
replace. This section summarizes a subset of results from a parallel,
more detailed study also under preparation for a separate
venue~\cite{doubledescentwork}.

\subsection{Experimental design}

Four condition arms isolating specific mechanisms, crossed with a 9-point
DNN width/depth capacity ladder (648K to tens of millions of parameters),
with architecture family (DeepFM) and data held fixed:

\begin{table}[h]
\centering
\caption{Double-descent condition arms.}
\label{tab:ddarms}
\small
\setlength{\tabcolsep}{4pt}
\begin{tabular}{@{}llll@{}}
\toprule
Arm & Regularization & Early stop & Label noise \\
\midrule
\texttt{clean\_prod} & production & on (patience 15) & 0\% \\
\texttt{nostop} & production & off & 0\% \\
\texttt{lowreg} & none & off & 0\% \\
\texttt{noisy} & none & off & 15\% \\
\bottomrule
\end{tabular}
\end{table}

\texttt{noisy} matches Nakkiran et al.'s sharpest-interpolation-peak
configuration.

\subsection{Results}

Table~\ref{tab:ddresults} distills the 4-arm $\times$ 9-capacity-point
grid to the migration-relevant comparison.

\begin{table*}[t]
\centering
\caption{Double-descent capacity scan, distilled to the migration-relevant
  comparison (full 36-configuration grid in Section~\ref{sec:dd}).}
\label{tab:ddresults}
\small
\setlength{\tabcolsep}{4pt}
\begin{tabular}{@{}lp{0.22\linewidth}rrrr@{}}
\toprule
Configuration & DNN shape (params) & CL & ER & GZ & Test PR-AUC \\
\midrule
Legacy (CatBoost) & --- & 0.4985 & 0.5310 & 0.5676 & --- \\
DeepFM baseline, 40ep & (2048,1024,512,256), 147.8M & 0.4678 & 0.5283 & 0.5158 & 0.6561 \\
Best size (\texttt{lowreg}), 40ep & (2048,1024,512,256), 17.2M$^{\dagger}$ & 0.4772 & 0.5508 & 0.5244 & \textbf{0.6921} \\
\textbf{Best noise (\texttt{noisy})}, 40ep & (2048,1024,512,256), 17.2M$^{\dagger}$ & \textbf{0.5003} & 0.5528 & 0.5368 & 0.6365 \\
200ep check (\texttt{noisy}) & (4096,2048), 37.1M, 200ep & 0.4800 & 0.5544 & 0.5222 & 0.6504 \\
\bottomrule
\end{tabular}
\end{table*}

$^{\dagger}$Same DNN tower shape as the DeepFM baseline model, but without
the $\sim$130M-parameter dense-projection layer that the DeepFM baseline's
architecture routes tabular features through --- so this is a genuinely
smaller model, not just the same model trained differently; it is 72$\times$
smaller in total parameters than the DeepFM baseline.

\textbf{Headline finding.} Nearly every one of the 36 configurations in the
full grid beats the DeepFM baseline model on test PR-AUC and CL F1, at a
fraction of the parameter count --- capacity alone was not the source of
the DeepFM baseline's ranking quality. Model size and regularization choice
are two separate levers: holding the same 17.2M-parameter DNN shape fixed,
switching from low regularization alone to noise injection moves CL from
0.4772 to \textbf{0.5003} --- the only configuration across the full grid
that beats CatBoost's own CL (0.4985), and it also clears CatBoost's ER
(0.5528 vs.\ 0.5310). GZ falls short of CatBoost (0.5368 vs.\ 0.5676) in
this configuration, so \textbf{no configuration clears CatBoost on all
three contact metrics simultaneously}; we report this honestly rather than
rounding up. The 200-epoch check answers a different question --- whether
training the same regularization recipe far longer at a larger capacity
point recovers more than the model-size/noise-injection search already
found --- and it does not (Section~\ref{sec:dd-epochwise}).

\textbf{Noise injection as a design lever, not just a diagnostic.} Label
noise acting as an implicit regularizer that \emph{improves} recall at
small capacity, rather than hurting it, is a usable knob for this
system's regularization schedule, not purely a research diagnostic --- we
adopt it, alongside the smaller model size above, as a design lever for
the next production iteration.

\subsection{Does the recovery appear at longer training horizons? A null result}
\label{sec:dd-epochwise}

The capacity scan above varies model size at a fixed epoch budget. As an
independent check, we asked whether the classical double-descent
\emph{epoch-wise} recovery appears if we simply train longer at fixed
capacity instead. We trained the \texttt{noisy} arm for 200 epochs at a
$(4096,2048)$-shape model (37.07M params) --- an order of magnitude
longer than any other run in this paper.

The run shows a single early peak in validation PR-AUC (epoch 16, 0.7428)
followed by monotone decay through the full 200-epoch budget (epoch 199,
0.2902), with no dip$\rightarrow$rise$\rightarrow$dip recovery at any
point (test PR-AUC 0.6504, CL 0.4800, ER 0.5544, GZ 0.5222). We report
this as a genuine null result, not a caveat to explain away: at this
capacity point, training longer is not itself a lever for recovering the
interpolation-onset quality loss --- the recovery this paper does observe
(Table~\ref{tab:ddresults}) comes from \emph{capacity and regularization}
choice, not from training duration.

\section{Conditioning on Transcript Text: Attention Mechanisms Across Architectures}
\label{sec:text}

\subsection{Motivation}

As noted in Section~\ref{sec:intro}, conversation chunks are not equally
informative, so fixed, content-agnostic pooling dilutes or discards signal
as length grows. Attention-based pooling~\cite{vaswani2017attention}
fixes this by learning a per-chunk relevance weight instead. Because a
full DeepFM run is expensive to iterate on, we first ran a lightweight
exploration on the smaller, cheaper two-tower architecture; once that
identified attention over the chunk sequence as the highest-leverage
lever, we carried it into DeepFM.

\begin{figure}[h]
\centering
\begin{tikzpicture}[
    font=\scriptsize,
    utt/.style={draw, rounded corners, minimum width=0.72cm, minimum height=0.38cm, fill=gray!12},
    op/.style={draw, rounded corners, fill=gray!30, minimum height=0.38cm, align=center, text width=3.55cm},
    vec/.style={draw, rounded corners, fill=black!75, text=white, minimum height=0.38cm, align=center},
    lbl/.style={font=\scriptsize\bfseries},
]

\node[lbl, align=left, text width=3.7cm] (titleA) at (0,3.5) {(a) DeepFM: shared-query attention pooling};

\node[utt] (a1) at (-1.35,2.85) {$u_1$};
\node[utt] (a2) at (-0.45,2.85) {$u_2$};
\node[utt] (a3) at (0.45,2.85)  {$u_3$};
\node[utt] (a4) at (1.35,2.85)  {$u_4$};

\node[op] (ascore) at (0,2.15) {shared learned query $\rightarrow$ scorer $\rightarrow$ masked softmax};
\node[vec] (apool) at (0,1.35) {pooled $\mathbf{c}_u$};
\node[align=center, font=\scriptsize] (adnn) at (0,0.75) {DNN tower input\\(concatenated with tabular)};

\foreach \n in {a1,a2,a3,a4} { \draw[->] (\n) -- (ascore); }
\draw[->] (ascore) -- (apool);
\draw[->] (apool) -- (adnn);

\draw[dashed] (0,0.35) -- (0,0.15);

\node[lbl, align=left, text width=3.7cm] (titleB) at (0,-0.15) {(b) Two-tower: candidate-conditioned target attention};

\node[utt] (b1) at (-1.35,-0.85) {$u_1$};
\node[utt] (b2) at (-0.45,-0.85) {$u_2$};
\node[utt] (b3) at (0.45,-0.85)  {$u_3$};
\node[utt] (b4) at (1.35,-0.85)  {$u_4$};

\node[vec, minimum width=1.3cm] (item) at (-2.55,-1.65) {item $p$};
\node[op, text width=3.2cm] (bscore) at (0.35,-1.65) {query $=$ [item $p$ embed.; target-enc.\ feats] $\rightarrow$ scorer};
\node[vec] (bpool) at (0.15,-2.45) {$\Delta_p(\text{transcript})$};
\node[align=center, font=\scriptsize] (bres) at (0.15,-3.05) {additive correction to\\zero-init.\ base logit};

\foreach \n in {b1,b2,b3,b4} { \draw[->] (\n) -- (bscore); }
\draw[->] (item) -- (bscore);
\draw[->] (bscore) -- (bpool);
\draw[->] (bpool) -- (bres);

\end{tikzpicture}
\caption{Both mechanisms replace fixed pooling with a learned, per-chunk
  relevance weight over the same per-utterance embedding sequence
  ($u_1,\ldots,u_4$): (a) DeepFM uses one shared query, producing a single
  pooled vector fed only to the DNN tower; (b) two-tower's query is
  conditioned on the specific candidate product $p$, so different
  candidates attend differently to the same call, producing an additive
  correction to the cold-safe base logit.}
\label{fig:attention-mechanisms}
\end{figure}
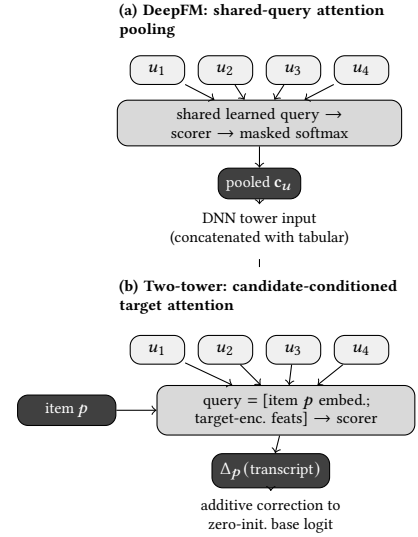

\subsection{Two-Tower: Residual Base Path plus Target-Attention Correction}
\label{sec:twotower}

This subsection reports the two-tower line of work: a cold-safe residual
architecture and the ablation campaign that improved it (25 trained
variants; we report the milestones). All numbers use the
fixed-population CL/ER/GZ evaluator of Section~\ref{sec:metrics}, so
they are directly comparable to the CatBoost baseline and to every
DeepFM configuration in this paper. Alongside CL/ER/GZ we report
\textbf{Top-1}, the argmax recommendation accuracy over the same
population (the best configuration below, TT-full, trains on a 58\%
chunk-stride subsample rather than 100\% of pair-rows).

\subsubsection{Architecture: a cold-safe base path plus a zero-initialized
  correction path}
\label{sec:tt-arch}

The design constraint is the one that motivates this paper's hard
no-regression requirement (Section~\ref{sec:setup}): the model must remain
trustworthy at \emph{call start}, when no transcript signal exists, while
exploiting transcript signal aggressively once it arrives. We meet it
structurally rather than by loss weighting alone:

\begin{itemize}
  \item \textbf{Base path (cold-safe).} A standard two-tower
    scorer~\cite{huang2013dssm}: a user/context tower over tabular,
    behavioral, and target-encoded features, and an item tower over the
    candidate-product representation, fused to a base logit. This path
    receives no transcript-sequence input, so its behavior is defined even
    at \texttt{chunk\_num}$=0$.
  \item \textbf{Correction path (transcript-aware, zero-initialized).} A
    second head reads the conversation-chunk sequence and outputs an
    additive correction: $\mathrm{logit} = \mathrm{logit}_{\text{base}} +
    \Delta(\text{transcript})$. The correction head's final layer is
    \textbf{zero-initialized}, so at initialization the model is exactly
    its cold-safe base --- the transcript path can only \emph{earn} its
    influence during training. Every later architectural addition was
    required to preserve this contract (verified by a unit test:
    $\mathrm{logit} \equiv \mathrm{logit}_{\text{base}}$ at init).
  \item \textbf{Target attention in the correction path.} The
    highest-impact addition (Section~\ref{sec:tt-experiments}): DIN-style
    target attention~\cite{zhou2018din}, where the attention query over
    transcript chunks is the concatenation of the candidate product's
    learned embedding and the row's per-product target-encoded features.
    Each candidate product thus attends to \emph{different} parts of the
    same call --- the natural fit for pairwise-binary scoring, where the
    same context is scored once per candidate.
  \item \textbf{Cold-focused loss shaping.} Two levers applied together:
    (i) a cold sample-weight rebalance that upweights
    \texttt{chunk\_num}$=0$ rows by $1 + \alpha\,(r - 1)$, where $r$ is
    the warm-to-cold row ratio and $\alpha{=}0.5$, and (ii) an auxiliary
    loss on the \emph{base} logit
    computed on cold rows only ($\lambda{=}0.5$), which trains the base
    path directly rather than letting the correction path absorb all
    gradient.
\end{itemize}

\subsubsection{Ablation campaign: what moved the metrics}
\label{sec:tt-experiments}

Table~\ref{tab:textablation} shows the milestone
configurations; each successive row was adopted only after replication,
since run-to-run seed noise on this data is $\approx$1pp on CL/Top-1. The
two highest-impact experiments of the campaign:

\textbf{(1) Target attention in the correction path: +3.3pp Top-1.} Adding
the DIN-style attention head lifted Top-1 from 0.5822 to 0.6152 --- the
largest single-lever gain of the campaign, replicated in two of three
independent runs --- while leaving ER intact. Pooling that is \emph{not}
conditioned on the candidate (uniform or recency-weighted, our v14 base)
leaves this headroom because a support call typically discusses several
needs and a single pooled summary blurs them.

\textbf{(2) Cold loss levers: recovering call-start quality.} Aggressive
transcript modeling tends to erode call-start (CL) quality --- gradient
concentrates in the correction path. The two cold levers above restored
CL to parity with the base architecture (0.4546 vs.\ 0.4575) while
retaining the attention gains, producing the best-balance configuration
(TT-full: best ER and GZ of the campaign). Global reweighting alone was
not sufficient.

TT-attn differs from TT-base by the target-attention head plus cold
levers; TT-full additionally trains on the 58\% chunk-stride subsample.
The line beats the legacy model decisively once transcript signal
exists (ER +2.8pp, GZ +5.2pp for TT-full) and trails only at call start
(CL $-$4.4pp), where no transcript is available (see
Table~\ref{tab:textablation} below for the full comparison).

\subsection{DeepFM: Attention Pooling}
\label{sec:text-arch}

The two-tower exploration above identified attention over the chunk
sequence as the highest-leverage lever for conditioning on transcript
text. We carry that lever into DeepFM directly, implementing it as an
attention-pooling layer over the same per-utterance embedding sequence,
and compare it against the text-representation baselines already
established for DeepFM.

\textbf{Per-utterance embedding store.} Dense text representations, using
\texttt{nomic-embed-text-v1.5} (768 dims), are computed once per
\emph{utterance} instead of once per \texttt{chunk\_num}, cached in a
chronologically-ordered flat store per contact; a row for
$(\mathrm{contact}, \mathrm{chunk\_num})$ reads a prefix slice of that
store, removing the $6.19\times$ redundant embedding calls naive
re-embedding would incur on our data.

\textbf{Pooling layer.} Given the variable-length sequence of per-utterance
embeddings available at a given chunk, a single learned-query attention
head scores each utterance --- a two-layer scorer
($768 \rightarrow 128$, $\tanh$, $128 \rightarrow 1$) shared across
positions --- and combines them via a masked softmax, with padding
positions scored at $-\infty$ before the softmax. Contacts with no
utterances yet available (\texttt{chunk\_num}$=0$, before any transcript
exists) receive an explicit zero vector rather than a softmax over an
empty/fully-masked sequence, which would otherwise produce undefined
(NaN) attention weights. The resulting pooled vector is concatenated onto
the DNN tower's input only, alongside the tabular features; it does not
enter the FM or linear components of the architecture.

\textbf{Integration.} The pooler is wrapped behind the same call interface
already used by the TF-IDF text projector it replaces, so the surrounding
training and evaluation code required no structural changes to accommodate
it --- only the text-representation module itself changed.

\subsubsection{Results}

\begin{table}[h]
\centering
\caption{Architecture and text-representation comparison, fixed-population CL/ER/GZ evaluator.}
\label{tab:textablation}
\small
\setlength{\tabcolsep}{4pt}
\begin{tabular}{@{}lrrr@{}}
\toprule
Configuration & CL & ER & GZ \\
\midrule
Legacy model (CatBoost) & \textbf{0.4985} & 0.5310 & 0.5676 \\
TT-base: residual two-tower & 0.4575 & 0.5537 & 0.6152 \\
TT-attn: + target attention & 0.4472 & 0.5537 & 0.6113 \\
TT-full: + cold levers, stride & 0.4546 & 0.5585 & 0.6195 \\
Tabular-only (DeepFM) & 0.3426 & 0.4362 & 0.4293 \\
DeepFM + TF-IDF/SVD + dense embeddings & 0.4678 & 0.5283 & 0.5158 \\
DeepFM + Attention pooling (ours) & 0.4867 & \textbf{0.5630} & \textbf{0.5943} \\
\bottomrule
\end{tabular}
\end{table}

The tabular-only vs.\ text-enabled comparison for DeepFM is strong
evidence that text signal matters, closing most of the CL/ER/GZ gap to
the CatBoost baseline (0.4985 / 0.5310 / 0.5676). Attention pooling goes
further: it is the first configuration in this paper to beat CatBoost
outright on ER and GZ (0.5630 vs.\ 0.5310, 0.5943 vs.\ 0.5676), within
1.2pp of it on CL. This is the strongest result in this paper and is
consistent with the loss/regularization picture in Section~\ref{sec:dd}.

\subsection{Synthesis}
\label{sec:arch}

Across both architectures, the deep model beats the legacy CatBoost model
outright only at ER and GZ --- the two evaluation points where a live
transcript actually exists --- and not at CL, where none does yet (closing
that gap is a negative-sampling/regularization lever, not an architecture
one; Section~\ref{sec:dd}). Two-tower gains ER/GZ via candidate-conditioned attention over the chunk
sequence but recovers only to near-parity with its own cold-safe base at
call start (Section~\ref{sec:twotower}); DeepFM gains ER/GZ via the same
underlying mechanism applied to a structurally different base
architecture, and comes closer to CatBoost at CL (within 1.2pp). This
ER/GZ effect appearing twice, via two different attention mechanisms on
two different architectures, is why we treat conditioning on transcript
text --- not the choice between DeepFM and two-tower --- as the factor
that justifies this migration, even though the two otherwise trade off
differently (Table~\ref{tab:textablation}): neither architecture
dominates the other, with two-tower's TT-full ahead on GZ (0.6195 vs.\
0.5943 for DeepFM's attention-pooling configuration, +5.2pp over
CatBoost) --- the metric measured at the point a recommendation is most
likely to fire --- which we attribute to the two-tower model's
target-attention correction path (Section~\ref{sec:tt-arch}), which
conditions transcript pooling on the candidate product and therefore
benefits most exactly when the transcript is longest.

\section{Inference-Time Challenges and Optimizations}
\label{sec:inference}

Moving from a boosted-tree ensemble to a deep recommender changes where
time is spent at serving time. We benchmarked both systems' real inference
paths end to end (base model only, no calibration) on the same machine,
using the production CatBoost artifact and the attention-pooling
checkpoint from Table~\ref{tab:textablation}. This section reports what
we found and the one change --- batching --- that closed most of the gap.

\textbf{Both systems pay a text featurization cost before the model ever
  runs --- far larger for CatBoost.} CatBoost's model call itself is
cheap (2.4ms); nearly all of its 641ms end-to-end cost is TF-IDF
vectorization. DeepFM pays the same kind of cost at a much smaller scale,
via a small transformer (nomic-embed) text embedding:

\begin{table}[h]
\centering
\caption{DeepFM's single-request cost by step (average, cold request).}
\label{tab:deepfmbreakdown}
\small
\setlength{\tabcolsep}{4pt}
\begin{tabular}{@{}lrrr@{}}
\toprule
Step & CPU & GPU & \% of CPU total \\
\midrule
Text embedding (nomic-embed) & 59.8ms & 34.3ms & 48\% \\
Tabular featurization & 12.5ms & 22.6ms & 10\% \\
DeepFM forward pass & 52.3ms & 56.4ms & 42\% \\
\midrule
Total (end to end) & 124.5ms & 113.3ms & 100\% \\
\bottomrule
\end{tabular}
\end{table}

On CPU, the text embedding is actually the larger of DeepFM's two costs
--- text featurization is the single most expensive part of the request
on both systems, more expensive than either model's own scoring logic.
On a single request with no batching, DeepFM is already faster end to end
than CatBoost (124ms/113ms vs.\ 641ms), but that is the wrong comparison
for a production system handling many requests at once --- and that is
where the picture reverses.

\textbf{Batching is the lever that matters.} A tree ensemble's
\texttt{predict()} call already scores many rows in one pass essentially
for free, so CatBoost's throughput scales almost linearly with batch size
(9{,}954 req/s at batch 64). A neural network gets the same benefit only
if the serving code actually batches requests together --- naively
handling each on its own thread makes things \emph{worse} under load
(throughput drops past 4 concurrent requests, as internal and thread
parallelism fight for the same cores). Correctly batched, DeepFM's
throughput rises from 17 req/s (batch 1) to 323 on CPU and 575 on GPU
(batch 64) --- $19\times$/$34\times$ on the same weights. This closes
most, not all, of the gap: CatBoost is intrinsically cheaper per row.
Batching brings DeepFM from unusable to workable at production load; it
does not make DeepFM cheaper than CatBoost outright.

\textbf{Batching lowers cost per request, not the latency of any one
  request --- larger batches make per-request latency \emph{worse}, not
  better.} A request has to wait for either enough others to fill a batch
or a wait timer to expire before it is scored at all. We measured this
with a request-coalescing server (holds requests up to 8ms, or until 64
arrive) under sustained, realistic (Poisson) arrival traffic. The
throughput numbers below assume enough concurrent contacts to fill a
batch; at low volume, real throughput falls back toward single-request
numbers:

\begin{table}[h]
\centering
\caption{Single-request and batch-64 serving cost by model (no
  coalescing). CL/ER/GZ/PR-AUC are properties of the model weights.}
\label{tab:single-request}
\scriptsize
\setlength{\tabcolsep}{2.5pt}
\begin{tabular}{@{}llrrrrrr@{}}
\toprule
Model & Setup & P50 & Thpt. & CL & ER & GZ & PR-AUC \\
\midrule
CatBoost, fp32 & batch 64 & --- & 9954/s & 0.4985 & 0.5310 & 0.5676 & --- \\
CatBoost, fp32 & single req. & 641ms & --- & 0.4985 & 0.5310 & 0.5676 & --- \\
DeepFM, fp32 & CPU, single req. & 124ms & 323/s\textsuperscript{a} & 0.4799 & 0.5550 & 0.5918 & 0.7431 \\
DeepFM, fp32 & GPU, single req. & 113ms & 575/s\textsuperscript{a} & 0.4799 & 0.5550 & 0.5918 & 0.7431 \\
DeepFM, INT8 & CPU, single req. & 74ms & 591/s\textsuperscript{a} & 0.4788 & 0.5342 & 0.5989 & 0.7217 \\
\bottomrule
\end{tabular}
\vspace{1pt}
{\footnotesize\textsuperscript{a}Batch-64 throughput ceiling (req/s), not a coalescing-server measurement under real arrival traffic --- see the coalesced-load table below.}
\end{table}

\begin{table}[h]
\centering
\caption{DeepFM under a request-coalescing server (holds requests up to
  8ms, or until 64 arrive), sustained Poisson arrival traffic.}
\label{tab:coalescing}
\small
\setlength{\tabcolsep}{3pt}
\begin{tabular}{@{}lrrr@{}}
\toprule
Setup & P50 & P95 & Throughput \\
\midrule
CPU, coalesced, 150 req/s & 288--377ms & 455--747ms & 82--103 req/s \\
GPU, coalesced, 150 req/s & 116--121ms & 168--365ms & 99 req/s \\
GPU, coalesced, 500 req/s & 477ms & 648ms & 201 req/s \\
\bottomrule
\end{tabular}
\end{table}

At this moderate arrival rate, the CPU coalescing server's typical latency
(288--377ms) is \emph{worse} than scoring each request immediately with no
batching (124ms): not enough requests arrive within the wait window to
fill a batch, so requests mostly sit near the timeout. Batching only pays
for itself once arrival volume is high enough to fill batches quickly ---
the maximum-wait tunable trades worst-case latency against batch size,
and we have no default to recommend without production traffic data.

\textbf{Quantization is a modest, metric-dependent lever, not a free
  win.} INT8 dynamic quantization (32-bit $\to$ 8-bit weights, no
retraining) shrinks the model file by 75\% and raises batch-64 throughput
by 11\% (Table~\ref{tab:single-request}) --- real but well short of batching's
$19$--$34\times$ --- while shifting quality unevenly: CL and GZ are
essentially unchanged or slightly better, but Early-Reco F1 and Test
PR-AUC both drop $\sim$2 points. Given the modest gain and real cost, we
do not adopt it by default; if used, it should be weighed against the
metric a deployment cares about most.

\textbf{What this means for production.} DeepFM is servable at a cost and
speed usable in a live system, but only with batching in place, which
most inference frameworks do not do automatically; that cost is worth
the recommendation gains in Table~\ref{tab:textablation}
(Section~\ref{sec:text-arch}). A smaller or distilled text encoder is a
promising untested lever, since text embedding is nearly half of
DeepFM's CPU cost (Table~\ref{tab:deepfmbreakdown}); a GPU is needed
only for throughput, not to fit the model (1.4GB peak VRAM at batch 64).

\section{Discussion and Conclusion}

Across every architecture in this paper, deep models beat the legacy
CatBoost model outright only at ER and GZ --- the two evaluation points
where a live transcript actually exists. At CL, before any transcript
exists, CatBoost's tabular-only strength still shows, though negative
sampling and noise-injection regularization close most of that gap
(below) rather than leaving it fixed. We take this as the paper's
central finding: it is conditioning on transcript text --- not the
choice between DeepFM and two-tower --- that justifies this migration.
Everything else in this paper (negative sampling, contrastive loss,
double descent, architecture choice) is the practical work required to
reach that text-conditioned regime without regressing call-start quality.

\textbf{How the evidence connects.} Each stage of this migration built on
the one before it, and the results trace a single chain rather than a
list of independent experiments. Capping and structuring implicit
negatives (Section~\ref{sec:neg-taxonomy}), rather than synthesizing all
candidate negatives per chunk, is what made the pairwise-binary
formulation trainable at all; further variants on top of that substrate
(Section~\ref{sec:negsampling}) moved pairwise ranking but not reliably
the contact-level metrics. Building on that improved negative population, batch composition
and the contrastive objective proved additive rather than redundant with
it (Section~\ref{sec:contrastive-loss}): 50:50 group batching alone
recovered most of the ER/GZ gain, but adding the contrastive term on top
still beat batching alone on every contact-level metric (CL 0.4981, ER
0.5510, GZ 0.5354 vs.\ batching alone). Double descent then asked whether
the DeepFM baseline's remaining shortfall was a matter of capacity or
regularization: the scan (Section~\ref{sec:dd}) showed nearly every
smaller, better-regularized configuration beating the 147.8M-parameter
baseline, and switching from low regularization to noise injection at a
fixed 17.2M-parameter capacity moved CL from 0.4772 to 0.5003, the only
grid configuration that beats CatBoost's own CL. Finally, attention pooling over the
conversation transcript (Section~\ref{sec:text}) is the one intervention
in this paper, alongside two-tower's target attention, that beats CatBoost
outright rather than only narrowing the gap --- and it is the piece that
makes the earlier work legible: better negatives and a sharper objective
matter more once the model can actually condition on where in a long call
the relevant signal sits, rather than pooling it away uniformly.

\textbf{What this unlocks.} A pairwise-binary deep recommender that accepts
arbitrary item-side features, rather than a fixed set of output classes, is
the structural precondition for scoring products never seen during
training --- new SKUs and fast-growing catalogs a multiclass label space
cannot track without retraining. We have not measured that capability
directly here; evaluating it is a natural next step, not a new
architecture effort, given the techniques this paper validates.

\textbf{What we hope this contributes beyond our own deployment.} This is
a production migration case study, not a benchmark result: the levers
above held under a live, non-regression-constrained system --- the
setting most practitioners migrating off tree-based models actually work
in, not a benchmark-scale one. A practitioner facing the same migration
will not find a universal winner here, and we think that is the honest,
useful finding.

\section*{Ethical Considerations}

\textbf{Privacy.} Training data includes live customer-support conversation
transcripts. PII and other sensitive information are redacted from these
transcripts before they reach the featurization pipeline, both for the
data used in training and at inference time.

\textbf{Fairness.} A recommender trained on historical agent/contact
outcomes can encode and amplify historical bias in which products were
pitched to which customers. We have not run a fairness audit across
customer segments, nor applied any explicit bias-correction step, and we
state this as an open limitation rather than a solved problem. Retraining
on more recent data is a candidate partial mitigation --- it lets the model
track shifts in agent behavior over time --- but it does not, on its own,
remove bias already encoded in the historical outcomes the model learns
from.

\textbf{Safety and misuse.} A model that recommends discount/bundle offers
in real time during a live conversation has a plausible misuse path
(steering vulnerable customers toward unneeded purchases). Two guardrails
bound this risk downstream of the model: a human agent remains in the loop
and decides whether and how to raise a recommendation with the customer,
and the DynaPitch augmentation system~\cite{dynapitch} constrains what is
actually said --- it turns a recommended product into real-time pitching
text that is grounded in that product's actual features and benefits,
rather than allowing an arbitrary or unsubstantiated claim to reach the
customer. A direction
we intend to pursue is reframing the training objective around long-term
customer lifetime value rather than short-horizon conversion, so that the
model's incentives are aligned with sustained customer benefit rather than
with maximizing the immediate sale.

\textbf{Societal impact of automation.} This system is designed to augment,
not replace, human agent judgment: the model recommends, and a human agent
decides whether and how to raise it with the customer. We frame the system
accordingly throughout this paper, consistent with its actual design.

\bibliographystyle{ACM-Reference-Format}

\end{document}